\documentclass[letterpaper, 10 pt, conference]{ieeeconf}  
\pdfoutput=1

\IEEEoverridecommandlockouts                              

\usepackage{multicol}
\usepackage{graphicx}        
\usepackage{url}
\usepackage{color}
\usepackage{benktools}
\usepackage{indentfirst}
\usepackage{amsmath,graphics}
\usepackage{amssymb} 
\usepackage{nicefrac, multirow}
\usepackage{algpseudocode}
\usepackage{tikz}
\usepackage{pgfplots}
\pgfplotsset{compat=newest}
\usepackage{subfig}
\usepackage{textcomp}

\usetikzlibrary{backgrounds}
\numberwithin{equation}{section} 

\tikzstyle{every node}=[font=\small]

\usepackage{tabularx}
\usepackage{pbox}
\usepackage{ragged2e}
\usepackage{multirow}
\usepackage{floatrow}
\usepackage{booktabs}
\usepackage{makecell}
\newfloatcommand{capbtabbox}{table}[][\FBwidth]

\newcolumntype{L}[1]{>{\RaggedRight\hspace{0pt}}p{#1}}
\newcolumntype{R}[1]{>{\RaggedLeft\hspace{0pt}}p{#1}}
\newcolumntype{Z}{ >{\centering\arraybackslash}X }

\renewcommand{\phi}{\varphi}

\newcommand{\furlp}[1]{\colorbox{blue!10}{\href{run:/home/fulong/academia/library/papers/#1.pdf}{D}}}
\newcommand{\furlb}[1]{\colorbox{blue!10}{\href{run:/home/fulong/academia/library/books/#1.pdf}{D}}}

\newcommand{\bx}{\mathbf{x}}

\newcommand{\by}{\mathbf{y}}

\newcommand{\bt}{\mathbf{t}}

\newcommand{\mO}{\mathcal{O}}

\newcommand{\mB}{\mathcal{B}}
\newcommand{\mC}{\mathcal{C}}

\newcommand{\mU}{\mathcal{U}}
\newcommand{\mD}{\mathcal{D}}

\newcommand{\mM}{\mathcal{M}}

\usetikzlibrary{external}
\makeindex

\makeatletter
\let\NAT@parse\undefined
\makeatother
\usepackage[numbers]{natbib}

\title{Segmenting Unknown 3D Objects from Real Depth Images\\ using Mask R-CNN Trained on Synthetic Data}

\author{Michael Danielczuk$^{1}$, Matthew Matl$^{1}$, Saurabh Gupta$^{1}$,\\ Andrew Li$^{1}$, Andrew Lee$^{1}$, Jeffrey Mahler$^{1}$, Ken Goldberg$^{1,2}$
\thanks{$^{1}$Department of Electrical Engineering and Computer Science}%
\thanks{$^{2}$Department of Industrial Engineering and Operations Research}
\thanks{$^{1-2}$The AUTOLAB at UC Berkeley; Berkeley, CA 94720, USA}
\thanks{{\tt\small \{mdanielczuk, mmatl, sgupta, andrewyli, andrew\_lee, jmahler, goldberg\}@berkeley.edu}}%
}

\begin{document}
\maketitle

\begin{abstract}
The ability to segment unknown objects in depth images has potential to enhance robot skills in grasping and object tracking.
Recent computer vision research has demonstrated that Mask R-CNN can be trained to segment specific categories of objects in RGB images when massive hand-labeled datasets are available.
As generating these datasets is time-consuming, we instead train with synthetic depth images. Many robots now use depth sensors, and recent results suggest training on synthetic depth data can transfer successfully to the real world. 
We present a method for automated dataset generation and rapidly generate a synthetic training dataset of 50,000 depth images and 320,000 object masks using simulated heaps of 3D CAD models.
We train a variant of Mask R-CNN with domain randomization on the generated dataset to perform category-agnostic instance segmentation without any hand-labeled data and we evaluate the trained network, which we refer to as Synthetic Depth (SD) Mask R-CNN, on a set of real, high-resolution depth images of challenging, densely-cluttered bins containing objects with highly-varied geometry. SD Mask R-CNN outperforms point cloud clustering baselines by an absolute 15\% in Average Precision and 20\% in Average Recall on COCO benchmarks, and achieves performance levels similar to a Mask R-CNN trained on a massive, hand-labeled RGB dataset and fine-tuned on real images from the experimental setup. We deploy the model in an instance-specific grasping pipeline to demonstrate its usefulness in a robotics application.
Code, the synthetic training dataset, and supplementary material are available at \url{https://bit.ly/2letCuE}.
\end{abstract}

\section{Introduction}
\seclabel{introduction}
\textit{Category-agnostic instance segmentation}, or the ability to mask the pixels belonging to each individual object in a scene regardless of the object's class, has potential to enhance robotic perception pipelines for applications such as instance-specific grasping, where a target object must be identified and grasped among potentially unknown distractor objects in a cluttered environment.
For example, recent approaches to grasp planning on unknown objects have used deep learning to generate robot grasping policies from massive datasets of images, grasps, and reward labels ~\cite{mahlerdexnet2, ten2018using}.
While these methods are effective at generalizing across a wide variety of objects, they search for high-quality grasp affordances across an entire scene and do not distinguish between the objects they are grasping.
Nonetheless, these methods may be extended to plan grasps for a particular target object by constraining grasp planning to an object mask produced by category-agnostic instance segmentation.

\begin{figure}[t!]
    \centering
    \includegraphics[width=\linewidth]{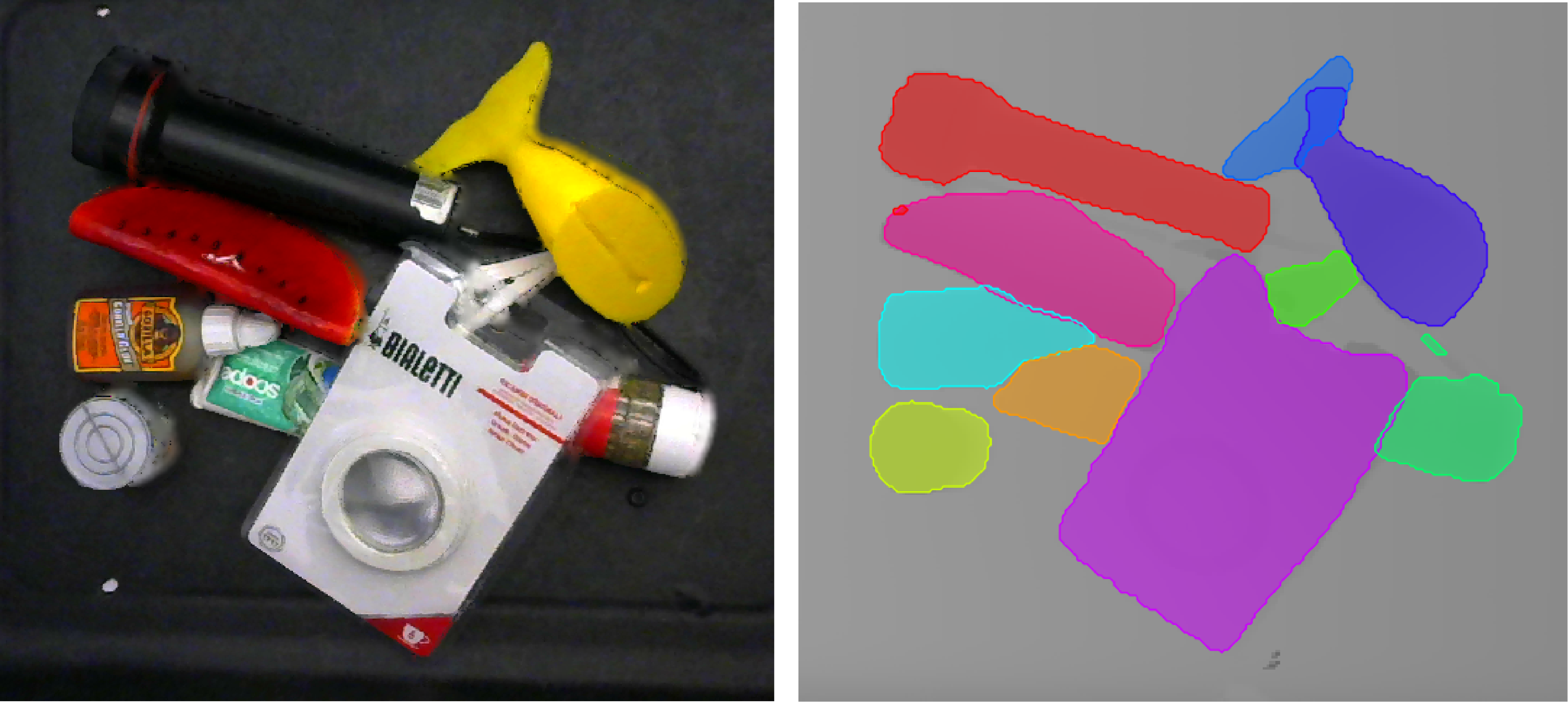}
    \caption{Color image (left) and depth image segmented by SD Mask RCNN (right) for a heap of objects. Despite clutter, occlusions, and complex geometries, SD Mask RCNN is able to correctly mask each of the objects.
    \vspace{-2ex}
    }
  \label{fig:real_dataset} 
\end{figure}

Object segmentation without prior models of the objects is difficult due to sensor noise and occlusions.
Computer vision techniques for generating category-agnostic object proposals~\cite{arbelaez2014multiscale, krahenbuhl2014geodesic} often oversegment and require secondary pruning steps to find a set of valid independent objects.
A variety of recent methods have demonstrated the ability to accurately segment RGB images into pre-defined semantic classes such as humans, bicycles, and cars by training deep neural networks on massive, hand-labeled datasets~\cite{long2015fully, he2017mask}.
These techniques require time-consuming human labeling to generate training data~\cite{lin2014microsoft}, and existing datasets consist of RGB images of natural scenes that are very different from the types of cluttered scenes commonly encountered in warehouses or fulfillment centers.
Adding new object classes or generating data for new types of environments requires additional manual labeling.
Thus, in robotics, pixel-wise object segmentation is often avoided~\cite{jang2017end, zeng2017multi} or used for a small number of object classes, where semantic segmentation networks~\cite{morrison2017cartman, schwarz2017nimbro} or predefined features~\cite{jonschkowski2016probabilistic} can be used.


To address these issues, we present a method and dataset for training Mask R-CNN~\cite{he2017mask}, a popular instance segmentation network, to perform category-agnostic instance segmentation on real depth images without training on hand-labeled data -- and, in fact, without training on real data at all.
We build on recent research which suggests that networks trained on synthetic depth images can transfer well from simulation to reality in some domains~\cite{johnson2017driving, mahler2017dex, ros2016synthia} and that depth cues can enhance instance segmentation in simulated images~\cite{shao2018clusternet}.  
To learn an instance segmentation network that transfers from simulation to reality, we propose to train on a large synthetic dataset of depth images with domain randomization~\cite{tobin2017domain} over a diverse set of 3D objects, camera poses, and camera intrinsic parameters.

This paper contributes:
\begin{enumerate}
   \item A method for rapidly generating a synthetic dataset of depth images and segmentation masks using domain randomization for robust transfer from simulation to reality.
   \item The Warehouse Instance Segmentation Dataset for Object Manipulation (WISDOM), a hybrid sim/real dataset designed for training and evaluating category-agnostic instance segmentation methods in the context of robotic bin picking.
   \item Synthetic Depth Mask R-CNN (SD Mask R-CNN), a Mask R-CNN adaptation designed to perform deep category-agnostic object instance segmentation on depth images, trained on WISDOM-Sim.
   \item Experiments evaluating the  sim-to-real generalization abilities of SD Mask R-CNN and performance benchmarks comparing it against a set of baseline instance segmentation methods.
\end{enumerate}

In an experimental evaluation on WISDOM-Real's high-resolution dataset, SD Mask R-CNN achieves significantly higher average precision and recall than baseline learning methods fine-tuned on WISDOM-Real training images, and also generalizes to a low-resolution sensor. We employ SD Mask R-CNN as part of an instance-specific grasping pipeline on an ABB YuMi bimanual industrial robot and find that it can increase success rate by 20\% over standard point cloud segmentation techniques.

\section{Related Work}
\seclabel{related-work}

This work builds on prior research in region proposal generation, 
neural architectures for image segmentation, and use of synthetic data 
for learning models in computer vision and robotics. The approach presented 
here is motivated and informed by robotic grasping, manipulation, and bin-picking tasks.

\paragraph{Box and Region Proposals} 
Early work in computer vision focused on using bottom-up cues for
generating box and region proposals in images \cite{arbelaez2014multiscale, krahenbuhl2014geodesic,alexe2012measuring, endres2010category, van2011segmentation}. 
Such techniques typically detect contours in images to obtain a hierarchical segmentation.
Regions from such a hierarchical segmentation are combined together and
used with low-level objectness cues to produce a list of regions that cover the 
objects present in the image. The focus of these techniques is on getting high recall, and 
the soup of output region proposals is used with a classifier to detect or segment
objects of interest \cite{carreira2012semantic}.

More recently, given advances in learning image representations, researchers have
used feature learning techniques (specifically CNN based models) to tackle this problem of producing 
bounding box proposals \cite{kuo2015deepbox, ren2015faster} and 
region proposals \cite{pinheiro2015learning, pinheiro2016learning}.
Unlike bottom-up segmentation methods, these techniques use data-driven methods 
to learn high-level semantic markers for proposing and classifying object segments.
Some of these learning-based region proposal techniques \cite{pinheiro2016learning} 
have built upon advances in models for image segmentation and use fine-grained 
information from early layers in CNNs \cite{long2015fully, hariharan2015hypercolumns}
to produce high quality regions.

While most work in computer vision has used RGB images to study these problems,
researchers have also studied similar problems with depth data. Once again there are
bottom-up techniques \cite{rusu2010semantic, rusu20113d, vo2015octree,
rabbani2006segmentation, gupta2013perceptual, gupta2014learning}
that use low-level geometry-based cues to come up with region proposals, as well as more recent
top-down learning-based techniques to produce proposals \cite{chen20183d}
in the form of image segments or 3D bounding boxes that contain objects in the scene.~\citet{shao2018clusternet} combined color and depth modalities, featurizing objects and clustering the features to produce object instance segmentation masks on simulated RGB-D images.

A parallel stream of work has tackled the problem of class-specific segmentation. Some 
of these works ignore object instances and study semantic segmentation \cite{long2015fully, 
pinheiro2016learning, garcia2017review, lin2017refinenet},
while others try to distinguish between instances 
\cite{he2017mask, hariharan2014simultaneous}.
Similar research has also been done in context of input from depth sensors
\cite{gupta2014learning, qi2017pointnet, ye2017depth, chen2015multi, wang2018sgpn}. 


\paragraph{Synthetic Data for Training Models}
Our research is related to a number of recent efforts for 
rapidly acquiring large training datasets containing image and 
ground truth masks with limited or no human labeling.
The most natural way is to augment training with synthetic 
color and depth images collected in simulation.
This idea has been explored extensively for training semantic segmentation networks 
for autonomous driving~\cite{johnson2017driving, ros2016synthia} and for estimating human and object pose \cite{shotton2011real, su2015render}.
Another approach is to use self-supervision to increase 
training dataset size by first hand-aligning 3D models 
to images with easy-to-use interfaces~\cite{marion2018pipeline}
or algorithmically matching a set of 3D CAD models 
to initial RGB-D images~\cite{zeng2017multi},
and then projecting each 3D model into a larger 
set of images from camera viewpoints with known 6-DOF poses. In comparison, we generate synthetic training datasets for
category-agnostic object segmentation in a robot bin picking domain.

\paragraph{Robotics Applications}
Segmentation methods have been applied extensively to grasping target objects, most notably in the Amazon Robotics Challenge (ARC).
Many classical grasping pipelines consisted of an alignment phase, in which 3D CAD models or scans are matched to RGB-D point clouds, and an indexing phase, in which precomputed grasps are executed given the estimated object pose~\cite{ciocarlie2014towards}.
In the 2015 ARC, the winning team followed a similar strategy, using a histogram backprojection method to segment objects from shelves and point cloud heuristics for grasp planning~\cite{eppner2016lessons}.
In 2016, many teams used deep learning to segment objects for the alignment phase, training semantic segmentation networks with separate classes for each object instance on hand-labeled~\cite{schwarz2016rgb} or self-supervised datasets~\cite{zeng2017multi}.
Team ACRV, the winners of the 2017 ARC, fine-tuned RefineNet to segment and classify 40 unique known objects in a bin, with a system to quickly learn new items with a semi-automated procedure~\cite{morrison2017cartman,milan2017semantic}.
In contrast, our method uses deep learning for category-agnostic segmentation, which can can be used to segment a wide variety of objects not seen in training. 

\section{Problem Statement} 
\seclabel{problem_statement}

We consider the problem of depth-based category-agnostic instance segmentation, or finding subsets of pixels corresponding to unique unknown objects in a single depth image.

To formalize category-agnostic instance segmentation, we use the following definitions:
\begin{enumerate}
    \item \textit{States:} Let $\bx = \{\mO_1, \hdots, \mO_m, \mB_1, \hdots, \mB_n, \mC\}$ be a ground truth state which contains (A) a set of $m$ foreground objects in the environment, (B) a set of $n$ background objects (e.g. bins, tables), and (C) a depth camera. Here, each object state $\mO_i$ or $\mB_j$ is defined by the object's geometry and 6-DOF pose, while the camera state $\mC$ is defined by its intrinsics matrix $K$ and its 6-DOF pose $(R, \bt) \in SE(3)$.
    \item \textit{Observations:} Let $\by \in \mathbb{R}^{H\times W}_+$ be a depth image observation of the state $\bx$ generated from $\mC$ with height $H$ and width $W$.
    Let the pixel space $\mU = [0, H-1] \times [0, W-1]$ be the set of all real-valued pixel coordinates in the depth image.
    \item \textit{Object Mask:} Let $\mathcal{M}_i \subseteq \mathcal{U}$ be a mask for foreground object $\mO_i$, or the set of pixels in $\by$ that were generated by the surface of $\mO_i$.
\end{enumerate}

Every state $\bx$ corresponds to a set of visible foreground object masks : $\mM = \{(\mM_i : \mM_i \neq \emptyset)\ \forall i \in \{1,\hdots,m\}\}$. 
The goal of category-agnostic object instance segmentation is to find $\mM$ given a depth image $\by$.


\section{Synthetic Dataset Generation Method} 
\seclabel{dataset-gen}

To efficiently learn category-agnostic instance segmentation, we generate a synthetic training dataset of $N$ paired depth images and ground truth object masks: $\mD = \left\{ (\by_k, \mM_k) \right\}_{k=1}^N$.
The proposed dataset generation method samples training examples using two distributions: a task-specific state distribution, $p(\bx)$, that randomizes over a diverse set of object geometries, object poses, and camera parameters; and an observation distribution, $p(\by | \bx)$, that models sensor operation and noise.

\begin{figure*}[t!]
    \centering
    \includegraphics[width=\linewidth]{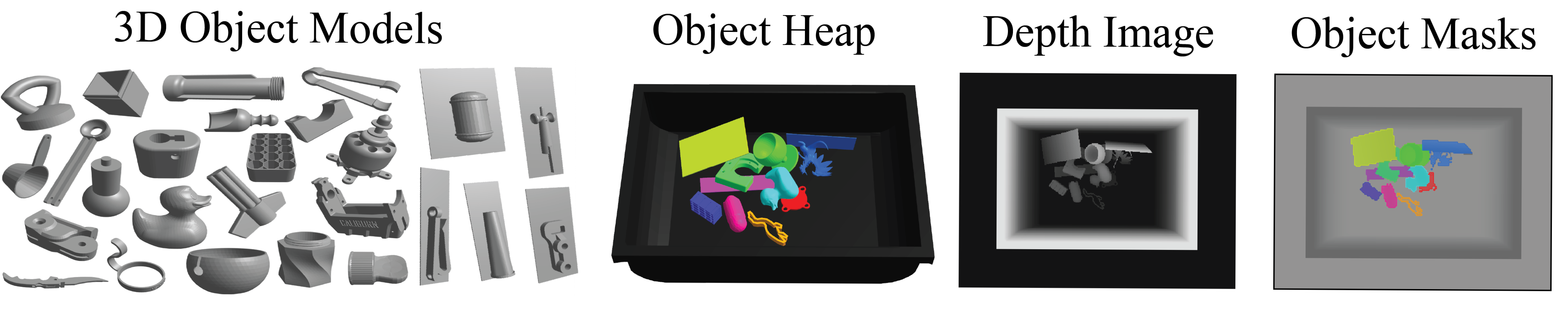}
    \caption{Dataset generation procedure for the WISDOM synthetic dataset. A subset of 3D CAD models from a training dataset of 1,600 objects are dropped into a virtual bin using dynamic simulation with pybullet. A virtual camera captures both a synthetic depth image of the scene and object segmasks based on the pixelwise projection of each unique 3D object. This process is repeated to generate 50,000 images.}
  \label{fig:synthetic_dataset} 
\end{figure*}

To sample a single datapoint, we first sample a state $\bx_k \sim p(\bx)$ using a dataset of 3D CAD models, dynamic simulation, and domain randomization~\cite{tobin2017domain} over the object states, camera intrinsic parameters, and camera pose for robust transfer from simulation to reality.
Next, we sample a synthetic depth image $\by_k \sim p(\by_k \mid \bx_k)$ using rendering.
Finally, we compute the visible object masks $\mM_j$ determining the set of pixels in the depth image with a corresponding 3D point on the surface of object $\mO_j$.
Specifically, we render a depth image of each object in isolation and add a pixel to the mask if it is within a threshold from the corresponding full-state depth image.


\section{WISDOM Dataset}

To test the effectiveness of this method, we generate the Warehouse Instance Segmentation Dataset for Object Manipulation (WISDOM), a hybrid sim/real dataset designed to train and test category-agnostic instance segmentation networks in a robotic bin-picking environment.
WISDOM includes WISDOM-Sim, a large synthetic dataset of depth images generated using the simulation pipeline, and WISDOM-Real, a set of hand-labeled real RGB-D images for evaluating performance in the real world.

\subsection{WISDOM-Sim}
For WISDOM-Sim, we consider an environment for robotic bin picking consisting of a table and a bin full of objects imaged with an overhead depth camera.
In general, $p(\bx)$ can be represented as a product over distributions on:
\begin{enumerate}
    \item \textit{Foreground and background object counts ($m$ and $n$)}: We draw $m$ from a Poisson distribution with mean $\lambda=7.5$, truncated to a maximum of 10. We set $n=2$ since we use two fixed background objects: a table and a bin.
    \item \textit{Background object states ($\{\mB_j\}_1^n$)}: We set the geometry and pose of the background objects to fixed values.
    \item \textit{Foreground object states ($\{\mO_j\}_1^m$)}: We sample the $m$ foreground objects uniformly from a dataset of 1,664 3D triangular mesh models from Thingiverse, including objects augmented with artificial cardboard backing to mimic common packages. Object poses are sampled by selecting a random pose above the bin from a uniform distribution, dropping each object into the bin one-by-one in pybullet dynamic simulation, and simulating until all objects come to rest~\cite{mahler2017learning}.
    \item \textit{Camera state ($\mC$)}: We sample camera poses uniformly at random from a bounded set of spherical coordinates above the bin. We sample intrinsic parameters uniformly at random from intervals centered on the parameters of a Photoneo PhoXi S industrial depth camera.
\end{enumerate}
\noindent Because the high-resolution depth sensor we use has very little white noise, we fix $p(\by | \bx)$ to simply perform perspective depth rendering using an OpenGL z-buffer.


We used these distributions to sample a dataset of 50,000 synthetic depth images containing 320,000 individual ground-truth segmasks.
Generating 50k datapoints took approximately 26 hours on a desktop with an Intel i7-6700 3.4 GHz CPU.
The synthetic images are broken into training and validation sets with an 80/20 split, where the split is both on images as well as objects (i.e. no objects appear in both the training and validation sets). The training set has 40,000 images of 1,280 unique objects, while the validation set contains 10,000 images of 320 unique objects.

\subsection{WISDOM-Real} 

\begin{figure}[h]
    \centering
    \includegraphics[width=\linewidth]{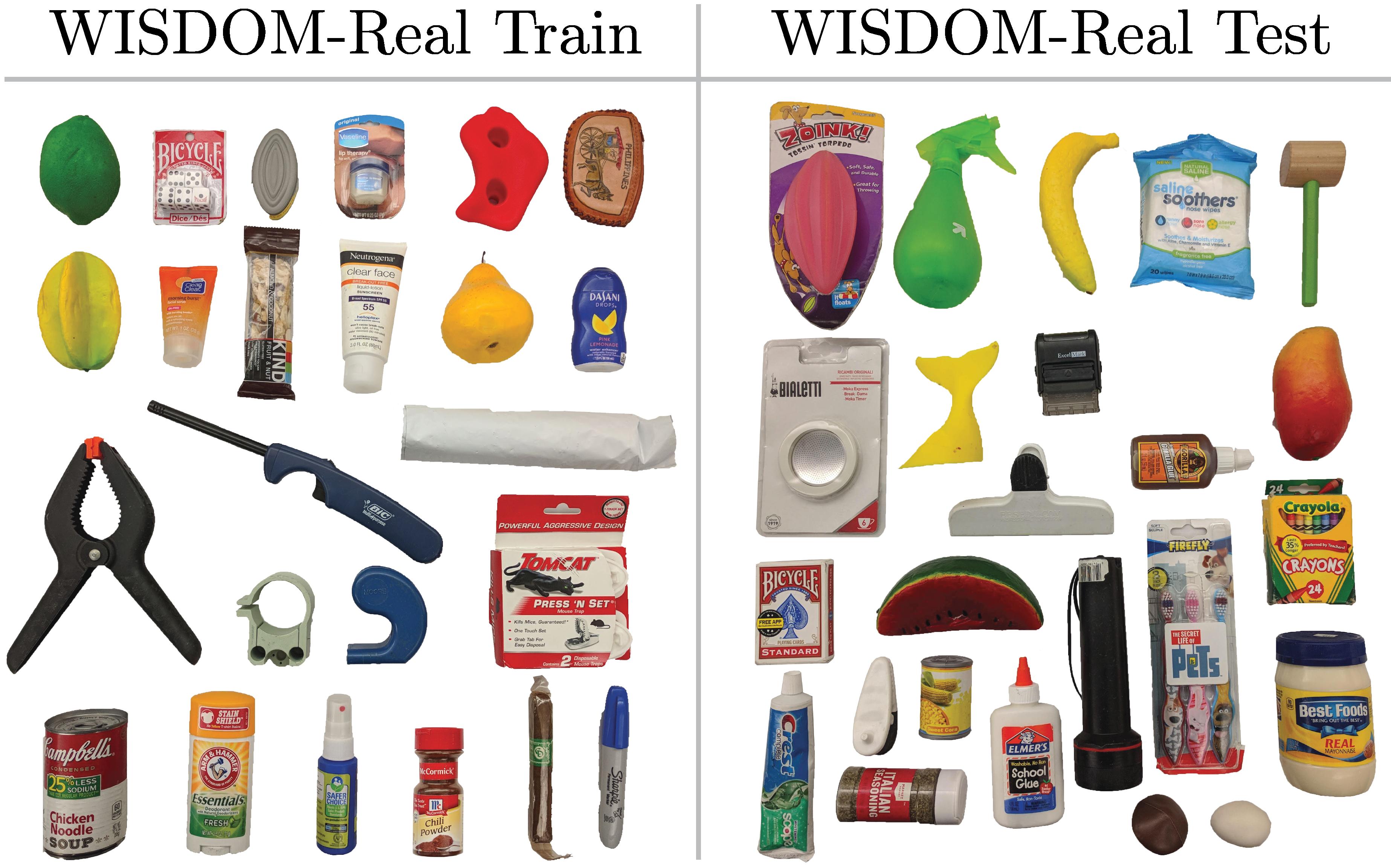}
    \caption{Objects included in the WISDOM-Real dataset. 25 objects were used for fine-tuning, while a separate set of 25 were held out for evaluation.}
  \label{fig:real_dataset} 
\end{figure}

\begin{figure}[h]
    \centering
    \includegraphics[width=0.9\linewidth]{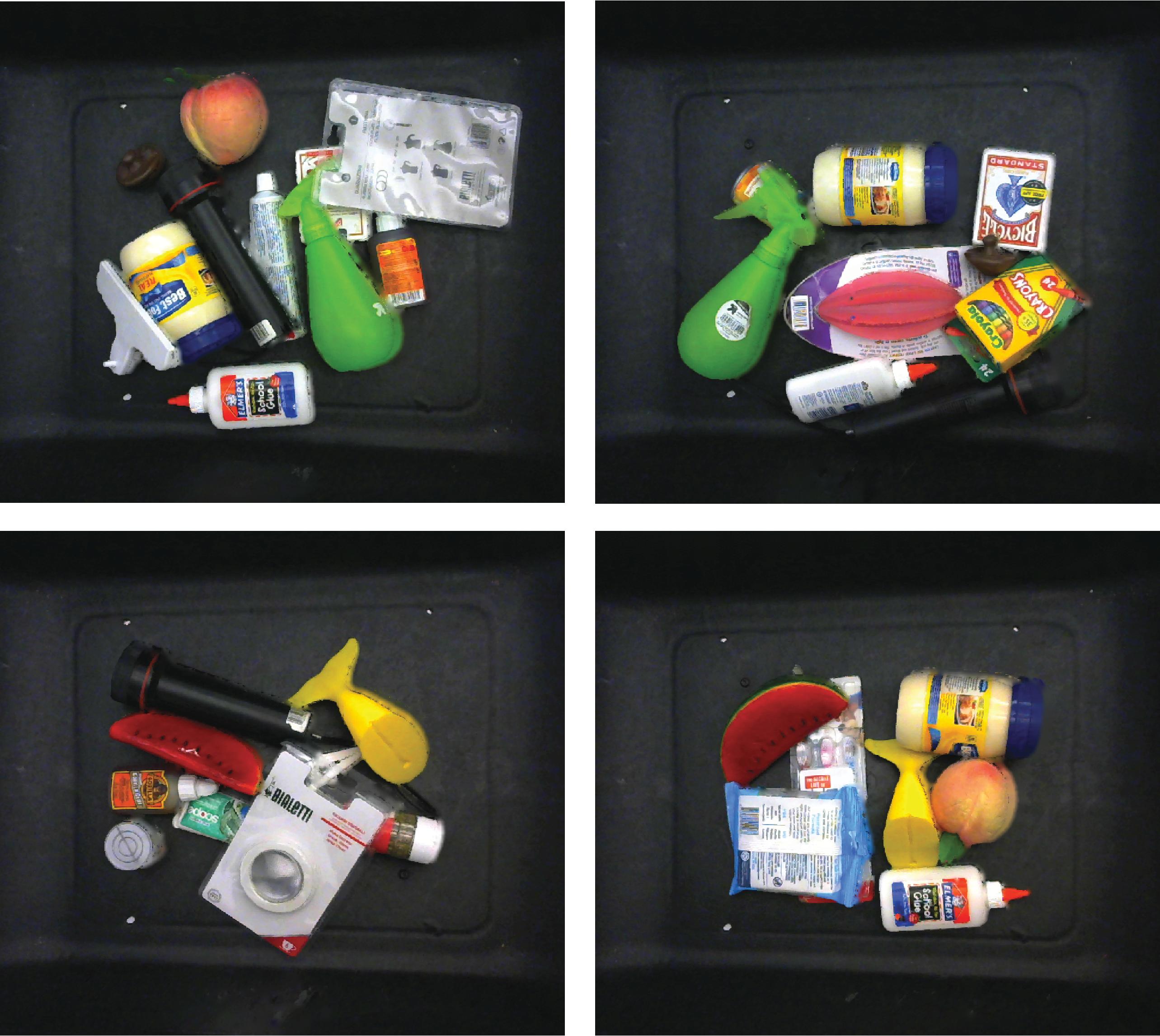}
    \caption{Example bins from the WISDOM-Real test set. The number of objects in each bin was chosen from a Poisson distribution with mean 5, with a minimum of two objects per bin. The highly-varied geometry and occlusions make these bins challenging to segment.}
  \label{fig:real_bins} 
\end{figure}

To evaluate the real-world performance of category-agnostic instance segmentation methods and their ability to generalize to novel objects across different types of depth sensors, we collected a hand-labeled dataset of real RGB-D images.
WISDOM-Real contains a total of 800 hand-labeled RGB-D images of cluttered bins, with 400 from both a high-resolution Photoneo PhoXi industrial sensor (1032x772 with 0.05 mm depth precision) and a low-resolution Primesense Carmine (640x480 with 1 mm depth precision). 
Missing depth values were filled in using fast inpainting~\cite{richard2001fast} with an averaging kernel.

The objects in these bins were sampled from a diverse set of 50 novel objects with highly-varied geometry, all of which are commonly found around the home (see Figure \ref{fig:real_dataset}), and have no corresponding CAD model in WISDOM-Sim.
This set of 50 objects was split randomly into even training and test sets.
The training set is reserved for learning methods that require real training data, while the test set is used to test generalization to novel objects.
We generated 100 bins containing objects from the training set and an additional 300 bins containing objects from the test set.
For each bin, a truncated Poisson distribution ($\lambda=5$) was used to determine the number of objects to be placed in the bin, and the objects were sampled uniformly at random from the appropriate subset.
The sampled objects were shaken together in a plastic box to randomize their poses and dumped into a large black bin with identical geometry to the bin used in WISDOM-Sim, and once the objects settled, each camera took an RGB-D image from above.
Sample bins are shown in Figure \ref{fig:real_bins}.

After dataset collection, all images were hand-labeled to identify unique object masks using the same tools used to label the COCO dataset~\cite{lin2014microsoft}.
We estimate that labeling the 800 real images took over 35 hours of effort due to time spent collecting on images, labeling object masks, and data cleaning.

\section{Synthetic Depth Mask R-CNN}
\seclabel{sdmrcnn}
To adapt Mask R-CNN to perform category-agnostic instance segmentation on depth images, we made several modifications:
\begin{enumerate}
\item We treat depth images as grayscale images and triplicate the depth values across three channels to match the input size of the original network.
\item We reduce the number of classes to two. Each proposed instance mask is classified as background or as a foreground object. Of these, only foreground masks are visualized.
\item We modify the network input to zero-pad the 512x384 pixel images in WISDOM-Sim to 512x512 images and set the region proposal network anchor scales and ratios to correspond to the 512x512 image size. 
\item For efficiency, we swapped out the ResNet 101 backbone with a smaller ResNet 35 backbone.
\item We set the mean pixel value to be the average pixel value of the simulated dataset.
\end{enumerate}

Training was based on Matterport's open-source Keras and TensorFlow implementation of Mask R-CNN from GitHub, which uses a ResNet 101 and FPN backbone~\cite{matterport2017mask}. This implementation closely follows the original Mask R-CNN paper in~\cite{he2017mask}. We made the modifications listed above and trained the network on WISDOM-Sim with an 80-20 train-val split for 60 epochs with a learning rate of 0.01, momentum of 0.9, and weight decay of 0.0001 on a Titan X GPU. On our setup, training took approximately 24 hours and a single forward pass took 105 ms (average of 600 trials). We call the final trained network a Synthetic Depth Mask R-CNN (SD Mask R-CNN).

\begin{table}[t!]
    \vspace{8pt}
    \setlength\tabcolsep{4pt}
    \begin{tabularx}{\linewidth}{*{1}{l}*{4}{Z}}
        \toprule
        & \multicolumn{2}{c}{High-Res} & \multicolumn{2}{c}{Low-Res} \\
        \thead{Method}  & \thead{AP} & \thead{AR} & \thead{AP} & \thead{AR} \\%
        \midrule
        Euclidean Clustering    & 0.324 & 0.467 & 0.183 & 0.317 \\
        Region Growing          & 0.349 & 0.574 & 0.180 & 0.346 \\
        FT Mask R-CNN (Depth) & 0.370 & 0.616 & 0.331 & 0.546\\
        FT Mask R-CNN (Color)  & 0.384 & 0.608 & \textit{\textbf{0.385}} & \textit{\textbf{0.613}} \\
        SD Mask R-CNN               & \textit{\textbf{0.516}} & \textit{\textbf{0.647}} & 0.356 & 0.465 \\
        \bottomrule
    \end{tabularx}
    \caption{Average precision and average recall (as defined by COCO benchmarks) on each dataset for each of the methods considered. SD Mask R-CNN is the highest performing method, even against Mask R-CNN pretrained on the COCO dataset and fine-tuned on real color and depth images from WISDOM-Real.}
    \label{tab:results}
\end{table}

\section{Experiments} 
\seclabel{experiments}

We compare performance of SD-Mask-R-CNN with several baseline methods for category-agnostic instance segmentation on RGB-D images.

\subsection{Baselines}
We use four baselines: two Point Cloud Library methods and two color-based Mask R-CNNs pre-trained on COCO and fine-tuned on WISDOM-Real images. For fine-tuning, the image shape and dataset-specific parameters such as mean pixel were set based on the dataset being trained on (e.g., either color or depth images).

\subsubsection{Point Cloud Library Baselines}
The Point Cloud Library, an open-source library for processing 3D data, provides several methods for segmenting point clouds~\cite{rusu20113d}. We used two of these methods: Euclidean clustering and region-growing segmentation. Euclidean clustering adds points to clusters based on the Euclidean distance between neighboring points. If a point is within a sphere of a set radius from its neighbor, then it is added to the cluster~\cite{rusu2010semantic}. Region-growing segmentation operates in a similar way to Euclidean clustering, but instead of considering Euclidean distance between neighboring points, it discriminates clusters based on the difference of angle between normal vectors and curvature~\cite{vo2015octree,rabbani2006segmentation}. We tuned the parameters of each method on the first ten images of the high-res and low-res WISDOM-Real training sets.

\subsubsection{Fine-Tuned Mask R-CNN Baselines}
As deep learning baselines, we used two variants of Mask R-CNN, one trained on color images and one trained on depth images triplicated across the color channels. Both these variants were pre-trained on RGB images from the COCO dataset and then fine-tuned using the 100 color or depth images from the WISDOM-Real high-res training set. All images were rescaled and padded to be 512 by 512 pixels, and the depth images were treated as grayscale images. Both implementations were fine-tuned on the 100 images for 10 epochs with a learning rate of 0.001.

\subsection{Benchmarks}
We compare the category-agnostic instance segmentation performance of all methods using the widely-used COCO instance segmentation benchmarks~\cite{lin2014microsoft}.
Of the metrics in the benchmark, we report average precision (AP) over ten IoU thresholds over a range from 0.50 to 0.95 with a step size of 0.05, and we report average recall (AR) given a maximum of 100 detections.
Averaging over several IoU thresholds rewards better localization from detectors, so we report this score as our main benchmark as opposed to simply the average precision for an IoU threshold of 0.50. All scores are for the segmentation mask IoU calculation. 

\begin{figure*}[t!]
    \vspace{8pt}
    \centering
    \includegraphics[width=\linewidth]{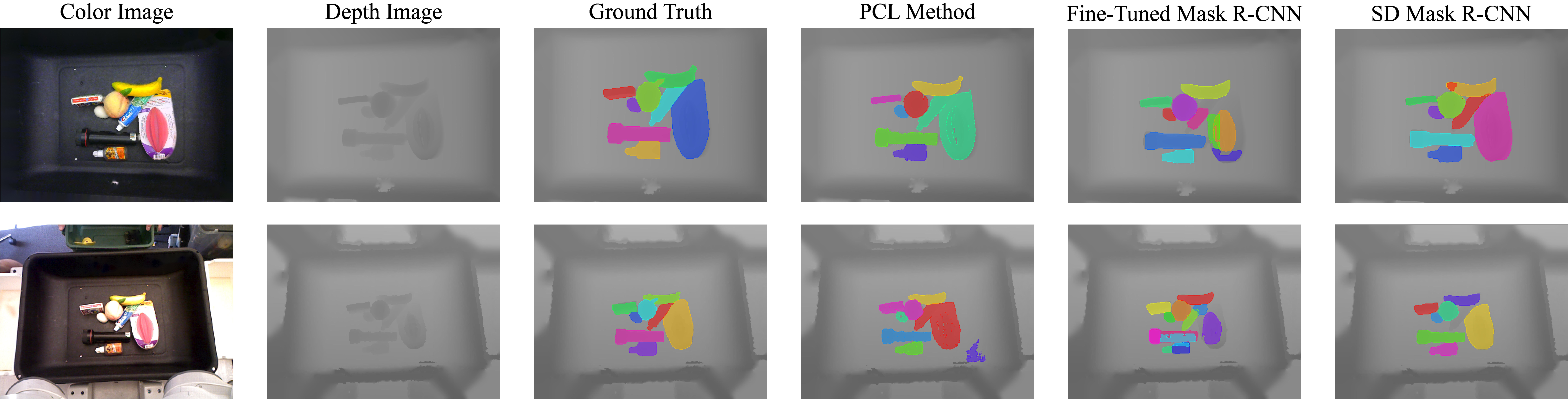}
    \caption{Images from both sensors with ground truth and object masks generated by each method. For the baseline methods, the better performing method was chosen for each scenario. While the baseline methods tend to undersegment (PCL) or oversegment (Fine-Tuned Mask R-CNN), SD Mask R-CNN segments the objects correctly.}
  \label{fig:comparison} 
\end{figure*}

\subsection{Performance}
We ran each of the methods on three test datasets: 2000 images from the WISDOM-Sim validation set and 300 real test images each from the Primesense and Phoxi cameras. All real test images were rescaled and padded to be 512 by 512 pixels.
The results are shown in Table~\ref{tab:results}, and full precision-recall curves for each dataset can be found in the supplemental file.
The SD Mask R-CNN network shows significant improvement over both the PCL baselines and the fine-tuned Mask R-CNN baselines, and is also robust to sensor noise.

An example of each method's performance on each of the real datasets can be seen in Figure~\ref{fig:comparison}.
The visualizations suggest that the PCL baselines tend to undersegment the scene and cluster nearby objects as a single object.
The fine-tuned Mask R-CNN implementations separate objects more effectively, but the color implementation may incorrectly predict multiple object segments on different colored pieces of the same object.
In contrast, the SD Mask R-CNN network can group parts of objects that may be slightly discontinuous in depth space, and is agnostic to color.
It is able to segment the scenes with high accuracy despite significant occlusion and variation in shape.
\tabref{results} also shows SD Mask R-CNN can perform similarly on low-res Primesense images, suggesting that the network can generalize to other camera intrinsics and poses.


\subsection{Robotics Application: Instance-Specific Grasping}
To demonstrate the usefulness of SD Mask R-CNN in a robotics task, we ran experiments utilizing category-agnostic instance segmentation as the first phase of an instance-specific grasping pipeline.
In this task, the goal is to identify and grasp a particular target object from a bin filled with other distractor objects.

We randomly selected a subset of ten objects from WISDOM-Real's test set and trained an instance classification network on ten RGB images of each object from a variety of poses by fine-tuning a VGG-16~\cite{simonyan2014very} classifier.
During each iteration, one object at random was chosen as the target and all of the objects were shaken, dumped into the black bin, and imaged with the Photoneo PhoXi depth sensor.
We then segmented the depth image using either SD Mask R-CNN or one of the baseline methods, colorized the mask using the corresponding RGB sensor image, and then labeled each mask with the pre-trained classifier.
The mask with the highest predicted probability of being the target was fed to a Dex-Net 3.0~\cite{mahler2017dex} policy for planning suction cup grasps, and the planned grasp was executed by an ABB YuMi equipped with a suction gripper.
Each iteration was considered a success if the target object was successfully grasped, lifted, and removed from the bin.

We measured performance on 50 iterations of this task each for three instance segmentation methods.
SD Mask R-CNN achieved a success rate of $74\%$, significantly higher than the PCL Euclidean clustering baseline ($56\%$).
Furthermore, SD Mask R-CNN had performance on par with a Mask R-CNN that was fine-tuned on 100 real color images ($78\%$).

\section{Discussion and Future Work}
\seclabel{discussion}
 We presented WISDOM, a dataset of images and object segmentation masks for the warehouse object manipulation environment, images that are currently unavailable in other major segmentation datasets. Training SD Mask R-CNN, an adaptation of Mask R-CNN, on synthetic depth images from WISDOM-Sim enables transfer to real images without expensive hand-labeling, suggesting that depth alone can encode segmentation cues. SD Mask R-CNN outperforms PCL segmentation methods and Mask R-CNN fine-tuned on real color and depth images for the object instance segmentation task, and can be used as part of a successful instance-specific grasping pipeline.

\figref{scaling} shows preliminary results of the effects of training image dataset size and training object dataset size on the performance of SD Mask R-CNN. We trained SD Mask R-CNN on random subsets of the training dataset with sizes $\{4k, 8k, 20k, 40k\}$ and on four $40k$ image datasets containing 100, 400, 800, and 1600 unique objects. Both AP and AR increase with image dataset size and number of unique objects used in training, although the image dataset size appears to have a stronger correlation with performance. These results suggest that performance might continue to improve with orders of magnitude more training data.

\begin{figure}[t!]
    \centering
    \includegraphics[width=\linewidth]{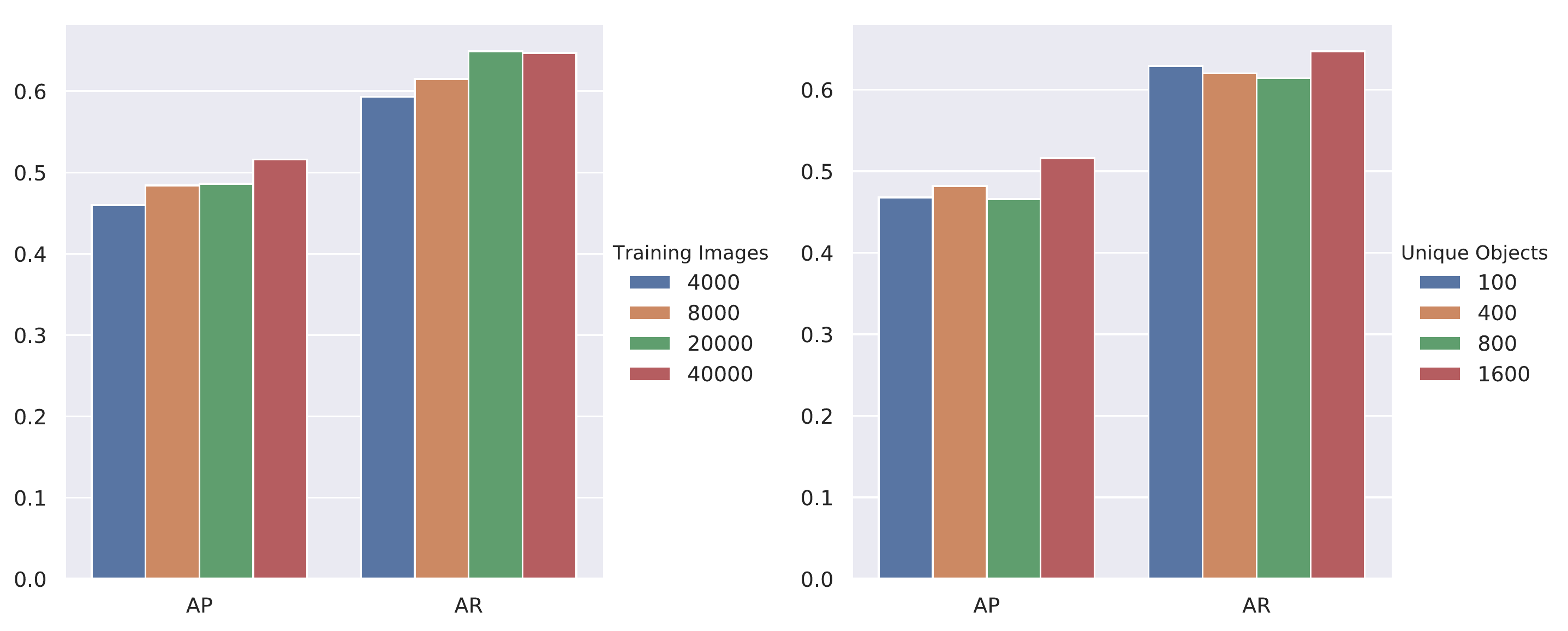}
    \caption{Effect of increasing synthetic training dataset size (left) and increasing the number of unique objects (right) on the performance of SD Mask R-CNN on the WISDOM-Real high-res test set. These results suggest that more data could continue to increase performance.}
  \figlabel{scaling}
\end{figure}
\section{Acknowledgments}
\footnotesize
This work was supported in part by a Google Cloud Focused Research Award for the Mechanical Search Project jointly to UC Berkeley's AUTOLAB and the Stanford Vision \& Learning Lab, in affiliation with the Berkeley AI Research (BAIR) Lab, Berkeley Deep Drive (BDD), the Real-Time Intelligent Secure Execution (RISE) Lab, and the CITRIS ``People and Robots" (CPAR) Initiative. The Authors were also supported by the Department of Defense (DoD) through the National Defense Science \& Engineering Graduate Fellowship (NDSEG) Program, the SAIL-Toyota Research initiative, the U.S. National Science Foundation under NRI Award IIS-1227536: Multilateral Manipulation by Human-Robot Collaborative Systems, Scalable Collaborative Human-Robot Learning (SCHooL) Project, the NSF National Robotics Initiative Award 1734633, and in part by donations from Siemens, Google, Amazon Robotics, Toyota Research Institute, Autodesk, ABB, Knapp, Loccioni, Honda, Intel, Comcast, Cisco, Hewlett-Packard and by equipment grants from PhotoNeo, and NVidia. Any opinions, findings, and conclusions or recommendations expressed in this material are those of the author(s) and do not necessarily reflect the views of the Sponsors. We thank our colleagues who provided helpful feedback, code, and suggestions, in particular Michael Laskey, Vishal Satish, Daniel Seita, and Ajay Tanwani.
\normalsize

\appendix
\section{Appendix}

\subsection{WISDOM Dataset Statistics}
\begin{figure*}[t!]
    \centering
    \includegraphics[width=\linewidth]{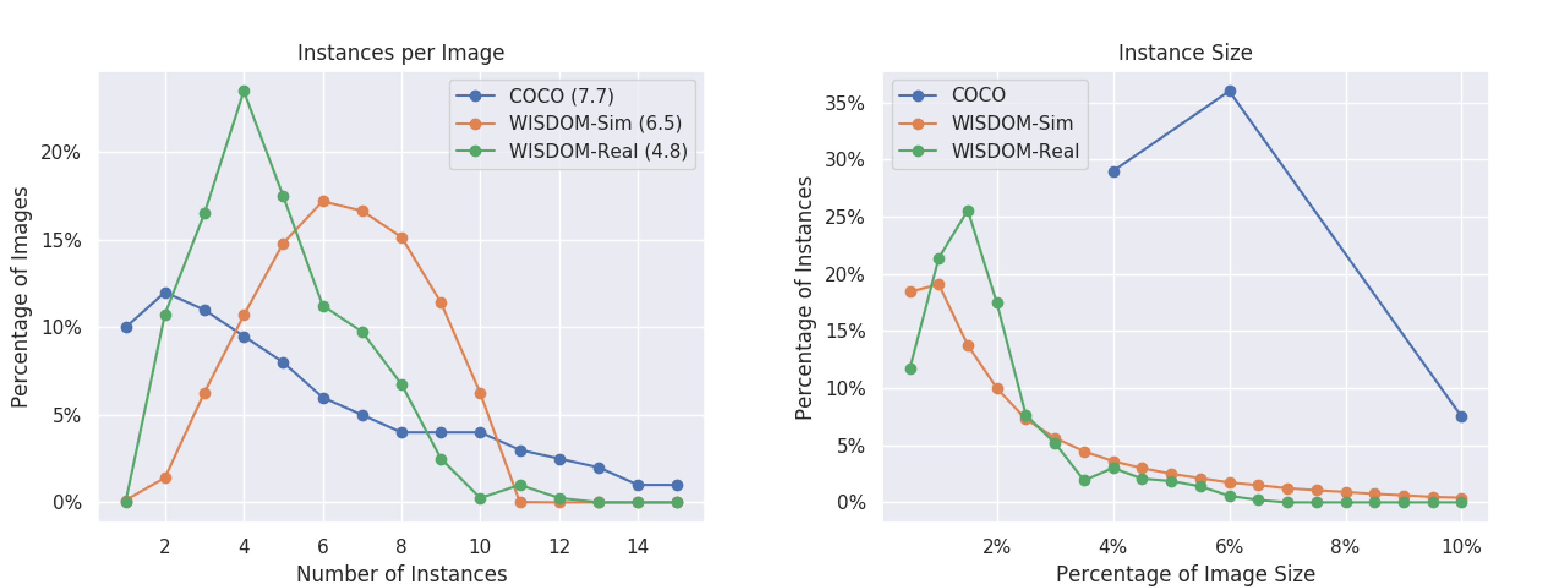}
    \caption{Distributions of the instances per image and instance size for the WISDOM dataset, with comparisons to the COCO dataset. Average number of instances are listed in parentheses next to each dataset. The number of instances and relative object size make this dataset more applicable to manipulation tasks.}
  \label{fig:WISDOM_stats}
\end{figure*}

The real dataset has 3849 total instances with an average of 4.8 object instances per image, fewer than the 7.7 instances per image in the Common Objects in Context (COCO) dataset and the 6.5 instances per image in WISDOM-Sim, but more instances per image than both ImageNet and PASCAL VOC (3.0 and 2.3, respectively). Additionally, it has many more instances that are close to, overlapping with, or occluding other instances, thus making it a more representative dataset for tasks such as bin picking in cluttered environments. Since it is designed for manipulation tasks, most of the objects are much smaller in area than in the COCO dataset, which aims to more evenly distribute instance areas. In the WISDOM dataset, instances take up only 2.28\% of total image area in the simulated images, 1.60\% of the total image area on average for the high-res images, and 0.94\% of the total image area for the low-res images. Figure~\ref{fig:WISDOM_stats} compares the distributions of these metrics to the COCO dataset.

\subsection{Precision-Recall Evaluation}
We performed additional experiments to analyze the precision-recall performance of SD Mask R-CNN along with the baseline methods for category-agnostic instance segmentation on RGB-D images: RGB object proposals~\cite{arbelaez2014multiscale, krahenbuhl2014geodesic}, cluster-based geometric segmentation methods from PCL~\cite{rusu20113d}, and Mask R-CNN fine-tuned for instance segmentation from the WISDOM-Real dataset.
We also include a variant of SD Mask R-CNN fine-tuned on real depth images from WISDOM-Real.
We evaluate performance using the widely-used COCO instance segmentation benchmarks~\cite{lin2014microsoft}.

The RGB object proposal baselines were based on two algorithms: Geodesic Object Proposals (GOP) \cite{krahenbuhl2014geodesic} and Multi-scale Combinatorial Grouping (MCG) \cite{arbelaez2014multiscale}. GOP identifies critical level sets in signed geodesic distance transforms of the original color images and generates object proposal masks based on these~\cite{krahenbuhl2014geodesic}. MCG employs combinatorial grouping of multi-scale segmentation masks and ranks object proposals in the image. For each of these methods, we take the top 100 detections. We then remove masks where less than half of the area of the mask overlaps with the foreground segmask of the image and apply non-max suppression with a threshold of $0.5$ Intersection-over-Union (IoU).

\figref{fig:pr_curves} shows precision-recall curves on three test datasets: 2000 images from the WISDOM-Sim validation set and 300 test images each from the Primesense and Phoxi cameras. 
Our learning-based method produces a ranked list of regions, that can be operated at different operating points.
Not only does SD Mask-RCNN achieve higher precision at the same operating point than PCL, it is able to achieve a much higher overall recall without any compromise in precision at the cost of only a handful of extra regions. 

\begin{figure*}[t!]
    \centering
    \includegraphics[width=\linewidth]{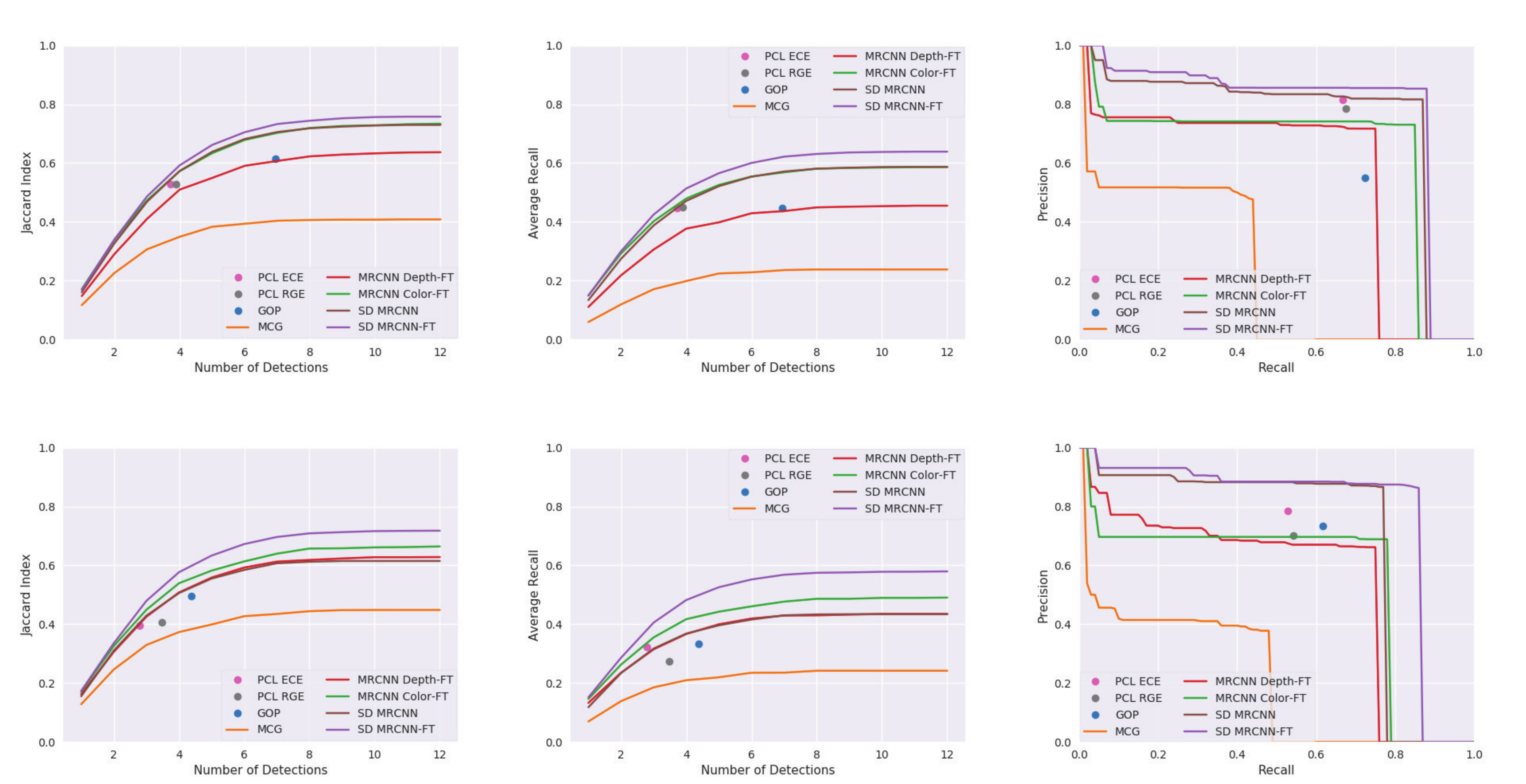}
    \caption{Average Jaccard Index, Average Recall, and Precision-Recall (at IoU = 0.5) curves for each method and the high-res (top row) and low-res (bottom row) dataset, using segmentation metrics. The fine-tuned SD Mask R-CNN implementation outperforms all baselines on both sensors in the WISDOM-real dataset. The precision-recall curves suggest that the dataset contains some hard instances that are unable to be recalled by any method. These instances are likely heavily occluded objects whose masks get merged with the adjacent mask or flat objects that cannot be distinguished from the bottom of the bin.}
  \figlabel{fig:pr_curves} 
\end{figure*}

We also evaluate the performance of SD Mask R-CNN and several baseline on the WISDOM-Sim test set.

\begin{table}[t!]
    \setlength\tabcolsep{4pt}
    \begin{tabularx}{\linewidth}{*{1}{l}*{2}{Z}}
        \toprule
        \thead{Method}  & \thead{AP} & \thead{AR} \\%
        \midrule
        Euclidean Clustering    & 0.161 & 0.252 \\
        Region Growing          & 0.172 & 0.274 \\
        SD Mask R-CNN           & \textit{\textbf{0.664}} & \textit{\textbf{0.748}} \\
        \bottomrule
    \end{tabularx}
    \caption{Average precision and average recall (as defined by COCO benchmarks) on the WISDOM-Sim dataset for the PCL baselines SD Mask R-CNN.}
    \label{tab:ap_sim}
\end{table}

\section{Details of Instance-Specific Grasping Experiment}
To evaluate the effectiveness of SD Mask R-CNN in a robotics task, we performed experiments in which we used segmentation as the first phase of an instance-specific grasping pipeline.
In the experiment, an ABB YuMi robot was presented a pile of ten known objects in a bin and instructed to grasp one specific target object from the bin using a suction gripper.
An attempt was considered successful if the robot lifted the target object out of the bin and successfully transported the object to a receptacle.

One approach to this problem is to collect real images of the items piled in the bin, labeling object masks in each image, and using that data to train or fine-tune a deep neural network for object classification and segmentation~\cite{morrison2017cartman, schwarz2017nimbro}.
However, that data collection process is time consuming and must be re-performed for new object sets, and training and fine-tuning a Mask R-CNN can take some time.
Instead, our experimental pipeline uses a class-agnostic instance segmentation method followed by a standard CNN classifier, which is easier to generate training data for and faster to train.

\begin{table*}[ht!]
    \setlength\tabcolsep{4pt}
    \begin{tabularx}{\linewidth}{*{1}{l}*{4}{Z}}
        \toprule
        \thead{Method}  & \thead{Success Rate ($\%$)}  & \thead{Prec. @ 0.5 ($\%$)} & \thead{$\#$ Corr. Targets} \\%
        \midrule
        Euclidean Clustering    & $56\pm14$ & $63\pm19$ & $35$ \\
        Fine-Tuned Mask R-CNN (Color)  & $78\pm11$ & $85\pm12$ & $44$ \\
        SD Mask R-CNN              & $74\pm12$ & $87\pm11$ & $39$\\  
        \bottomrule
    \end{tabularx}
    \caption{Results of semantic segmentation experiments, where success is defined as grasping and lifting the correct object. \textbf{(Success Rate)} Number of successful grasps of the correct object over 50 trials.  \textbf{(Prec. @ 0.5)} Success rate when the classifier was $>50\%$ certain that the selected segment was the target object. \textbf{(\# Corr. Targets)} Number of times the robot targeted the correct object out of 50 trials.}
    \label{tab:phys_results}
\end{table*}

To train the classifier, we collected ten RGB images of each target object in isolation.
Each image was masked and cropped automatically using depth data, and then each crop was augmented by randomly masking the crop with overlaid planes to simulate occlusions.
From the initial set of 100 images, we produced a dataset of 1,000 images with an 80-20 train-validation split.
We then used this dataset to fine-tune the last four layers of a VGG-16 network~\cite{simonyan2014very} pre-trained on Imagenet.
Fine-tuning the network for 20 epochs took less than two minutes on a Titan X GPU, and the only human intervention required
was capturing the initial object images.

Given a pre-trained classifier, the procedure used to execute instance-specific suction grasps was composed of three phases.
First, an RGB-D image was taken of the bin with the PhoXi and a class-agnostic instance segmentation method is used to detect object masks.
Then, the classifier was used to choose the mask that is most likely to belong to the target object.
Finally, a suction grasp was planned and executed by constraining grasps planned by Dex-Net 3.0 to the target object mask~\cite{mahler2017dex}.

We benchmarked SD Mask R-CNN on this pipeline against two segmentation methods.
First, we compared against the PCL Euclidean Clustering method to evaluate baseline performance.
Second, we compared with Mask R-CNN fine-tuned on the WISDOM-Real training dataset to evaluate whether SD Mask R-CNN is competitive with methods trained on real data.

Each segmentation method was tested in 50 independent trials.
Each trial involved shaking the ten objects in a box, pouring them into the bin, and allowing the system to select a target object uniformly at random to attempt to locate and grasp.
The results of these instance-specific grasping experiments are shown in Table \ref{tab:phys_results}.

SD Mask R-CNN outperforms the PCL baseline and achieves performance on par with Mask R-CNN fine-tuned on real data, despite the fact that SD Mask R-CNN was training on only synthetic data.
This suggests that high-quality instance segmentation can be achieved without expensive data collection from humans or self-supervision.
This could significantly reduce the effort needed to take advantage of object segmentation for new robotic tasks.

\bibliographystyle{IEEEtranN}
\bibliography{bibliography}  
\end{document}